\documentclass{bmvc2k}
\usepackage{lipsum}
\usepackage{framed,multirow}
\usepackage{booktabs}
\usepackage{url}
\usepackage{xcolor}

\usepackage{comment}

\usepackage{algorithm}
\usepackage[noend]{algorithmic}
\usepackage{eqparbox}

\usepackage{graphicx}
\usepackage{amsmath,amssymb} %
\usepackage{bm}
\usepackage{booktabs}
\usepackage{rotating,multirow}
\usepackage{wrapfig}

\usepackage{tikz}
\usepackage{transparent} %
\usepackage{calc} %
\usepackage[export]{adjustbox}
\usepackage{bbm} %

\def\ie{\emph{i.e}\bmvaOneDot} 
\def\cf{\emph{c.f}\bmvaOneDot}

\def\etal{\emph{et al}\bmvaOneDot}
\makeatother

\graphicspath{{./Images/}{./Figures/result/}{./Images/result/}}

\newcommand{\OLD}[1]{}

\newcommand{\da}{$\downarrow$}
\renewcommand{\bf}{\bfseries}

\newcommand{\pie}[1]{%
\begin{tikzpicture}
 \draw (0,0) circle (1ex);\fill (1ex,0) arc (0:#1:1ex) -- (0,0) -- cycle;
\end{tikzpicture}%
}

\newcommand{\hleft}{%
\begin{tikzpicture}
\draw[fill] (0, 0) -- (0ex, 0ex) arc(90:270:1ex) -- (0, 0);
\end{tikzpicture}%
}

\newcommand{\hright}{%
\begin{tikzpicture}
\draw[fill] (0, 0) -- (0ex, 0ex) arc(-90:90:1ex) -- (0, 0);
\end{tikzpicture}%
}

\newlength\myheight
\newlength\mydepth
\settototalheight\myheight{Xygp}
\newcommand*\inlinegraphics[1]{%
  \settototalheight\myheight{Xygp}%
  \settodepth\mydepth{Xygp}%
  \raisebox{-\mydepth}{\includegraphics[height=\myheight]{#1}}%
}

\definecolor{attention}{HTML}{d4aa00}

\title{Semantic Masks to Bounding Boxes and vice versa: disjoint multitask learning with hyperbolic contrastive loss} 
\title{Mask for Box and Box for Mask: disjoint multitask learning with hyperbolic contrastive loss} 
\title{Mask for Box and Box for Mask: multi-task partially supervised learning with hyperbolic contrastive loss} 
\title{Box for Mask and Mask for Box: weak losses for multi-task partially supervised learning} 

\addauthor{Hoàng-Ân Lê}{hoang-an.le@univ-ubs.fr}{}
\addauthor{Paul Berg}{paul.berg@irisa.fr}{}
\addauthor{Minh-Tan Pham}{minh-tan.pham@irisa.fr}{}

\addinstitution{
IRISA, Université Bretagne Sud, 
\\
UMR 6074, 56000 Vannes, France 
}

\runninghead{H.-Â. Lê, P. Berg, M.-T. Pham}{BoMBo: Box for Mask and Mask for Box}

\begin{document}

\maketitle

\begin{abstract}

Object detection and semantic segmentation are both scene understanding tasks yet they differ in data structure and information level. Object detection requires box coordinates for object instances while semantic segmentation requires pixel-wise class labels. Making use of one task's information to train the other would be beneficial for multi-task partially supervised learning where each training example is annotated only for a single task, having the potential to expand training sets with different-task datasets. This paper studies various weak losses for partially annotated data in combination with existing supervised losses. We propose \underline{Bo}x-for-Mask and \underline{M}ask-for-\underline{Bo}x strategies, and their combination BoMBo, to distil necessary information from one task annotations to train the other. Ablation studies and experimental results on VOC and COCO datasets show favorable results for the proposed idea. Source code and data splits can be found at \url{https://github.com/lhoangan/multas}.

\end{abstract}

\section{Introduction}
\label{sec:intro}
Multi-task learning is an active research area of computer vision, in which a shared model is used to optimize multiple targets together from the same input.
As such, the model is compelled to learn a shared representation from the task-specific information, thus becoming better generalized and achieving improved performance~\cite{Vandenhende2021}. Sharing models in the deep learning era also means reducing the number of parameters, or memory footprints, and increasing training and inference speed. As a result, efforts have been made to combine different image-level tasks~\cite{Rebuffi2018,Stoica2023zipit} or those with dense annotations~\cite{Zhang2018,Li2020KDMT,Saha2021,Li2022MTPSL,Vandenhende2021}.

As the improvement of multi-task learning resulted from the interrelationships obtained by co-training the tasks, the training examples are commonly assumed to be annotated for all the targets. This strong assumption would impede scalability, both in the number of tasks and the number of training examples, as it is expensive to maintain synchronization among the tasks, besides the multiplied annotating efforts. The situation becomes impossible for post hoc tasks requiring synchronized sensor reads, such as depth images.

Multi-task partially supervised learning (MTPSL) introduced by Li~\etal~\cite{Li2022MTPSL} relaxes the requirements as it allows each training example to associate only with one of the target tasks. Thus, adding a new task to an existing dataset would only involve adding new training examples annotated for the task without having to annotate all the existing images, effectively enlarging the training set. As such, it can also be seen as potentially augmenting the existing task with more examples if the task interrelationship can be learned from such settings. 

\begin{figure}[t]
    \centering
    \def\svgwidth{\textwidth}
    \tiny
    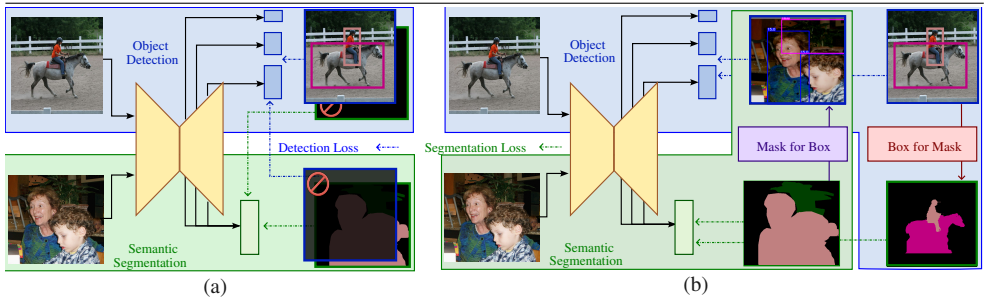
    \caption{Multi-task partially supervised learning with two tasks, object detection (blue) and semantic segmentation (green). (a) Each image is labeled for a single task, indicated by background colors, thus, can only train the respective head. (b) The proposed Mask-for-Box and Box-for-Mask modules allow training one task head from the other's ground truths.
         }
    \label{fig:MTL-general}
\end{figure}

This exploitation has been demonstrated to improve both tasks' performances by Lê and Pham~\cite{Le2023BMVC} with a naïve approach where each task is trained only using the respective annotations (Figure~\ref{fig:MTL-general}(a)).
Although experiments show favorable results thanks to the improved shared subnet, it is, generally, challenging as the task-specific part of the network cannot be trained with data only annotated for the other task, resulting in diminishing or even negative gain when the target spaces or data domains are too much different.

Inspired by these insights, we hypothesize that providing weak but relevant training signals for one task when the other task's ground truths are available would allow learning better joint representations of both tasks and improve performance. This paper evaluates various weakly-supervised methods of one task given the other's annotations and illustrates the idea with 2 semantic tasks, namely object detection and semantic segmentation as~\cite{Le2023BMVC}. These tasks, intuitively, are related to one another on the general target of scene understanding yet differ in data structure and information level. Semantic masks identify the object at each pixel yet lack the boundary between instances required for object detection; on the other hand, ground truth bounding boxes enclose the object instance but are short of pixel-wise information. Therefore, extracting the information that improves one task from the other's ground truth would provide more insights into their interrelationship.

Although pseudo-masks, in theory, can be generated from ground truth boxes using unsupervised methods like GrabCut~\cite{GrabCut}, and pseudo-boxes from ground truth masks' circumscribed rectangles, they are insufficient to effectively train a network. To that end, two modules, \underline{Bo}x-for-Mask and \underline{M}ask-for-\underline{Bo}x, and their combination BoMBo, are proposed, to refine and make use of one task's annotations and provide targets for training the other (Figure~\ref{fig:MTL-general}(b)). Our contributions, therefore, are:

\begin{itemize}
    \item We propose a refinement process to extract instance information from
    semantic masks guided by predicted bounding boxes;
    \item We evaluate various weak losses for training semantic segmentation
    from box annotations, and vice versa, with various network architectures;
    \item We present BoMBo including two modules Box-for-Mask and Mask-for-Box for
    multi-task partially supervised learning.
\end{itemize}

\section{Related work}
\label{sec:relatedwork}

\subsection{Multi-task partially supervised learning} %

Multi-task learning has been an interesting topic in the computer vision and machine learning community because of the potential efficiency and improvement over single-task setups. Notable research directions include sharing models~\cite{Bragman2019,Bruggemann2020,Gao2019,Liu2019mtan}, balancing task contribution~\cite{Guo2018MTL,Chen2020MTL,Liu2019mtan}, or task relationships~\cite{Zamir2020xtaskConsist,Li2022MTPSL,Lu2021taskology}, of which the dense-prediction tasks such as semantic segmentation, depth, and surface normal are prominent~\cite{Li2020KDMT,Li2022MTPSL,Vandenhende2021}.

Multi-task supervised learning~\cite{Rebuffi2018,Bragman2019,Bruggemann2020,Liu2019mtan,Guo2018MTL,Chen2020MTL} is, generally, expensive because of the multiplied efforts to maintain all-task synchronized annotations for the whole datasets.
Li~\etal~\cite{Li2022MTPSL} are the first to coin the term multi-task partially supervised learning and partially annotated data. Different from semi-supervised setups in which limited training data are assumed but still with all task annotations~\cite{Chen2020meanTeacher,Imran2020}, partially annotated data requires each training example to be annotated only for a single task. This paradigm, even with a naïve approach where the model is trained only on the available labels~\cite{Le2023BMVC}, has proven beneficial for improving performance. This paper focuses on scene understanding tasks initiated by Lê and Pham~\cite{Le2023BMVC},~\ie object detection and semantic segmentation, diverging from the dense-prediction task requirement of~\cite{Li2022MTPSL}.

\subsection{Learning object detection from semantic masks}

Training detection heads using semantic segmentation annotations can be considered as a weakly supervised problem in which semantic masks provide the object categories and weak localizing information. The common line of approach (\cf~\cite{Wang2023SSOD}) for weakly-supervised object detection consists of generating pseudo-labels using a duplicated model, called teacher, whose weights are not trained but updated from the main model by exponential moving average (EMA)~\cite{Zhou2021meanTeacher,Xu2021softTeacher,Liu2022ubteacher2,Li2022PseCo}.
The teacher and the student network are, therefore, fed with with same unlabeled inputs but with different augmentation levels.

To the best of our knowledge, no work relies on semantic segmentation ground truths to supervise object detection, intuitively, because pixel-wise semantic masks are more expensive than box annotations. In the context of this paper, semantic masks come from the assumption that a dataset designed for the semantic segmentation task is combined with another for object detection in a multi-task learning framework for the benefit of both tasks.

\subsection{Learning semantic segmentation from box annotations}

Learning semantic segmentation from box annotations is also a weakly supervised problem.
On the one hand, pseudo-masks can be generated from the ground truth bounding boxes with uncertainties using
unsupervised methods such as dense CRF~\cite{DenseCRF}, GrabCut~\cite{GrabCut}, and M\&G+~\cite{MCG}, On the other hand, the ground truth bounding boxes identify the enclosed objects' categories without delineation, thus can be used to confine predictions and treated as noisy masks by filling up enclosed areas with class labels.
Methods like~\cite{Song2019BCM,Song2023BCM,Kulharia2020Box2Seg} try to reduce the effect of erroneous gradients from wrong labelling by using filling-rates, as hard constrains~\cite{Song2019BCM,Song2023BCM} or regularization mechanism~\cite{Kulharia2020Box2Seg} while box-shape masks are used in attention mechanisms.

\section{Method}
\label{sec:method}

\begin{figure}[t]
    \centering
    \def\svgwidth{\textwidth}
    \scriptsize
    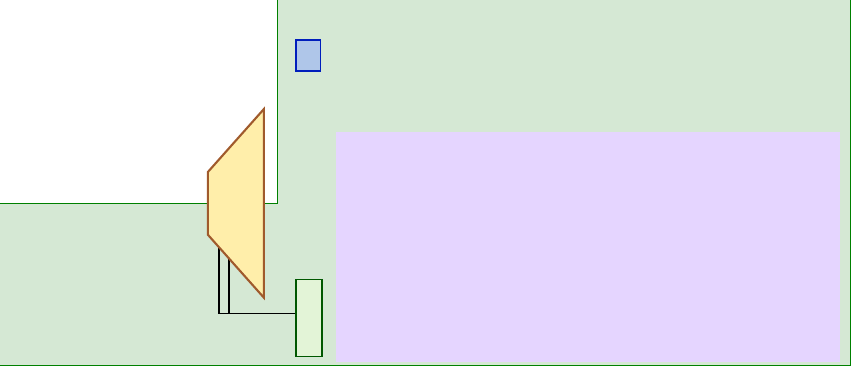
    \caption{The Mask-for-Box module uses predicted boxes to refine the circumscribed rectangles of the masks' connected components, by separating~\protect\inlinegraphics{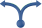} multi-instance masks, merging~\protect\inlinegraphics{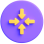} sub-instance masks, or using as ground truths~\protect\inlinegraphics{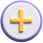}. The good~\protect\inlinegraphics{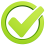} predicted boxes provide the instance cue while the wrong~\protect\inlinegraphics{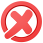} are to be removed. %
    }
    \label{fig:MTL-M4B}
\end{figure}

\begin{figure}[t]
    \centering
    \def\svgwidth{\textwidth}
    \scriptsize
    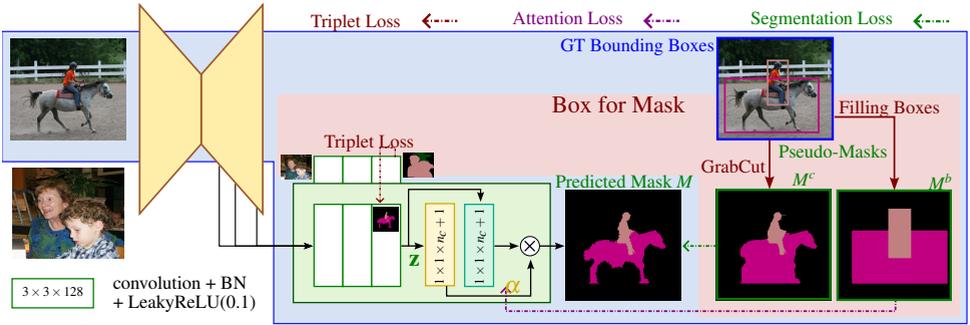
    \caption{The Box-for-Mask module generates pseudo-masks by filling the ground truth boxes with the same category and an unsupervised-learning method, like GrabCut~\cite{GrabCut}. The box-shaped pseudo-masks are used to train the \textcolor{attention}{attention map $\alpha$} while the other are the predicted masks. The triplet loss constrains the embeddings to follow those with annotations.
    }
    \label{fig:MTL-B4M}
\end{figure}

\subsection{Multi-task partially supervised learning}

In this paper, we follow the conventional multi-task learning architecture comprising a shared encoder network with backbone and neck subnet for extracting features and decoder heads outputting task-specific predictions.
The overview scheme is shown in Figure~\ref{fig:MTL-general}(a).
As each training example is annotated for only a single task, not all the losses can be optimized together.
Therefore, the network is fed with data annotated for one task after another, and the gradients are computed separately for each head but accumulated for the shared subnet. All parameters are updated once the data for both tasks have been passed through the network.
Object detection loss $\mathcal{L}_\text{det}$ includes the classification Focal Loss~\cite{Lin2017focal} and the localization Balanced L1 Loss~\cite{Pang2019}  while the semantic segmentation head uses the regular cross-entropy with softmax loss $\mathcal{L}_\text{seg}$. To make up for the lack of corresponding annotations of one task when training the other, the Mask-for-Box losses $\mathcal{L}_\text{M4B}$ and Box-for-Mask losses $\mathcal{L}_\text{B4M}$ are proposed to provide pseudo training signals, as shown in Figure~\ref{fig:MTL-general}(b). A detailed architectural scheme can be seen in the supplementary materials. The two task losses are balanced by the parameter $\lambda$ as follows:

\begin{equation}
    \label{eq:loss}
    \mathcal{L} =   \left(\mathcal{L}_\text{det} + \mathcal{L}_\text{M4B}\right) + \lambda
                    \left(\mathcal{L}_\text{seg} + \mathcal{L}_\text{B4M}\right)
\end{equation}

\subsection{Mask for Box}

The Mask-for-Box module is inspired by the observation that the circumscribed rectangles of the connected components in a semantic mask could be considered referenced boxes for training object detection. These boxes correctly identify the object's category, by definition, but might not the object's instances, thus introducing noise during training. On the other hand, predicted boxes of a well-trained network have learned to localize object instances, however, with less precision and may fail to recognize the correct object category. The refining idea, therefore, aims to take the best of both worlds by re-localizing the referenced boxes with the guidance of the predicted boxes' localization information. The process described in Figure~\ref{fig:MTL-M4B} shows the Mask-for-Box module applied to two input images. The circumscribed boxes of the connected components in the ground truth masks can be one of the three cases. They may (1) contain multiple object instances, which should be separated (indicated by~\inlinegraphics{split}), (2) cover only a part of an object, which should be merged with other boxes (\inlinegraphics{merge}), otherwise (3) they can be directly used for training (\inlinegraphics{add}). The referenced boxes are, first, matched with the predicted boxes following the regular object detection process. The predicted boxes not overlapping with any reference box of the same class are removed (indicated by~\inlinegraphics{rejected}) while those with confidence higher than $ \theta_1$ threshold are used for refinement (\inlinegraphics{accepted}). Reference boxes that cannot be matched to any predicted box are merged before being re-matched. For refinement, as it can be observed that a predicted box should have at least 2 sides touching those of a reference box, a refined box will have the two touching sides from the reference box and the other two from the predicted. Some examples of the refined boxes can be seen in Figure~\ref{fig:qualitativeM4B} while some failure cases and as well as the pseudo-code can be found in the supplementary materials.
Although the same supervised losses can be used for training with refined boxes,
only the localization loss is used following the experimental results in Table~\ref{tab:M4B-prelim}.

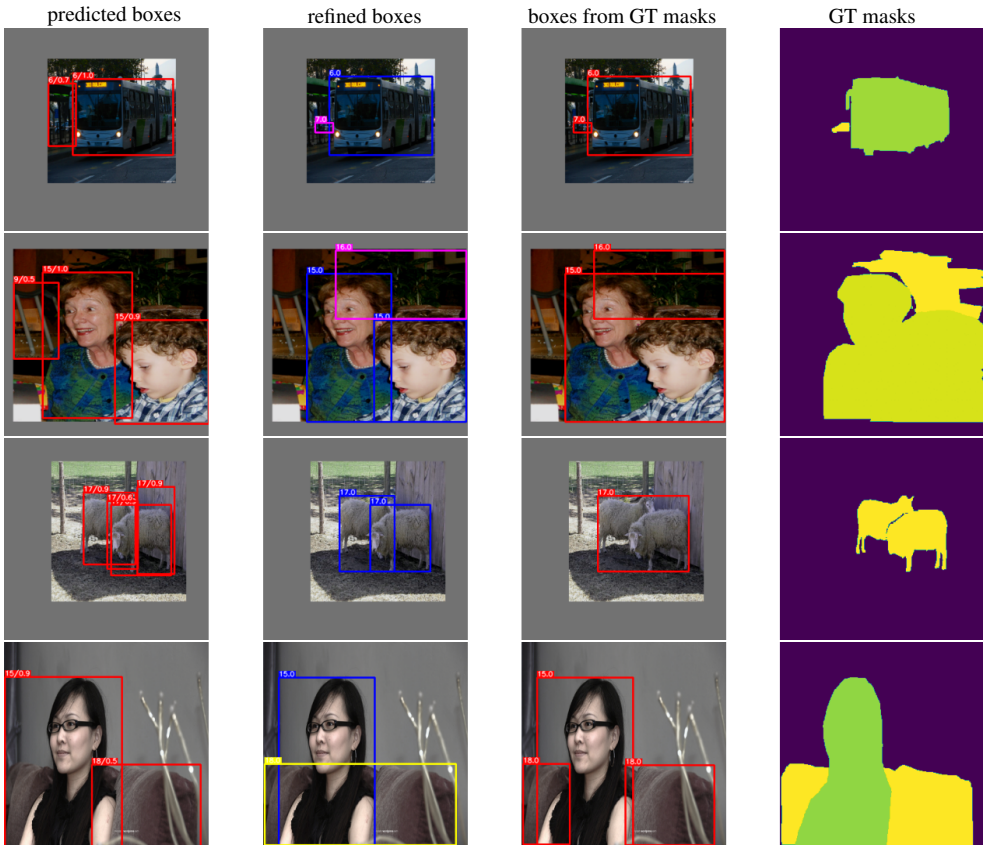
\begin{figure}[t]
  \centering
  \begin{tikzpicture}
    \node at (-5.00,1.5) {\footnotesize predicted boxes};
    \node at (-1.70,1.5) {\footnotesize refined boxes};
    \node at ( 1.70,1.5) {\footnotesize boxes from GT masks};
    \node at ( 5.00,1.5) {\footnotesize GT masks};
    \node at (0, 0) {
        \includegraphics[width=\linewidth]{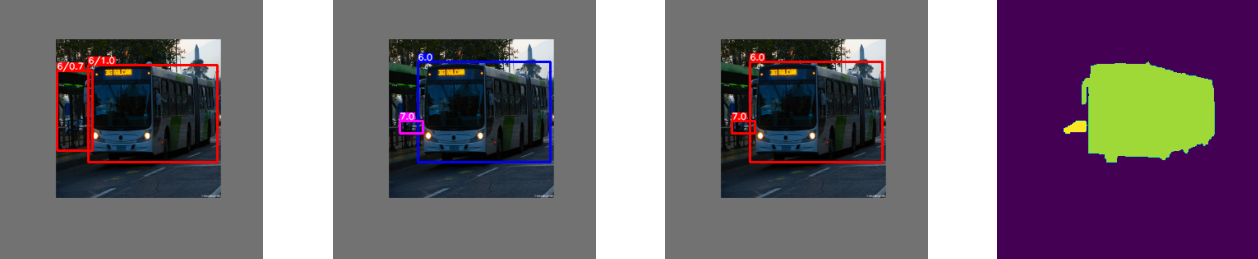} %
    };
    \node at (0, -2.7) {
        \includegraphics[width=\linewidth]{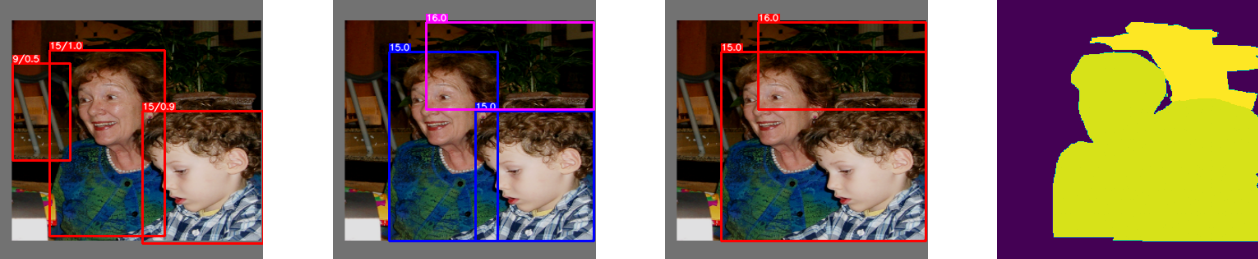} %
    };
    \node at (0, -5.4) {
        \includegraphics[width=\linewidth]{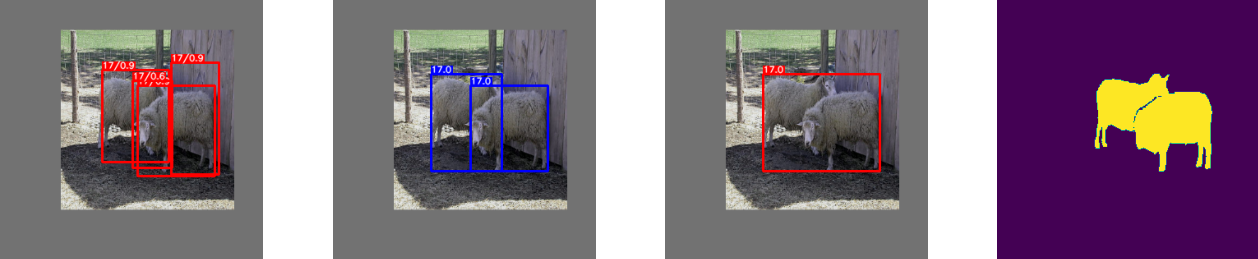} %
    };
    \node at (0, -8.1) {
        \includegraphics[width=\linewidth]{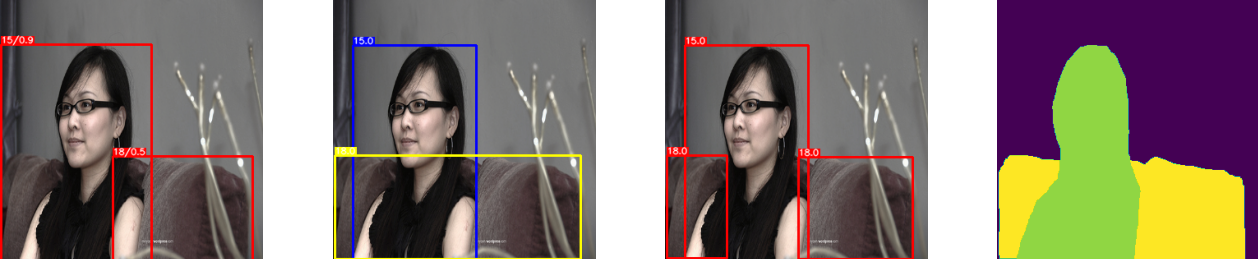} %
    };
  \end{tikzpicture}
  \vspace{-1em}
  \caption{
  Qualitative results of refined boxes with magenta indicate the adding, yellow merging boxes from ground truth masks, and blue for boxes from prediction.}
  \label{fig:qualitativeM4B}
\end{figure}

\subsection{Box for Mask}

To train the semantic segmentation head from ground truth bounding boxes, the Box-for-Mask module is proposed. The general scheme is illustrated in Figure~\ref{fig:MTL-B4M}. Let $\mathcal{C} = \{0, 1,\dots, n_c\}$ be the set of categories where $0$ denotes the background, and $N = w\times h$ the spatial dimension of the input image. Following previous work~\cite{Kulharia2020Box2Seg,Song2019BCM,Song2023BCM}, two types of pseudo-masks are generated from provided ground truth boxes, a box-shaped mask $M^b \in \mathcal{C}^{N}$ by filling a box-enclosed area with its category, with priority given to the smaller box, and a coarse pseudo mask $M^c \in \mathcal{C}^{N}$ using an unsupervised method such as GrabCut~\cite{GrabCut}, dense CRF~\cite{DenseCRF}, or MCG~\cite{MCG}, etc. As it can be observed that a ground truth box provides the upper limit of the object's extent, the coarse mask $M^c$ is filtered by $M^b$ so that $M^c_i = 0, \forall i: M^b_i = 0$, where $M^b_i = 0$ denoting a pixel $i$ in the box-shaped mask not enclosed in any ground truth box, thus surely a background pixel.

\vspace{-1em}
\paragraph{Semantic segmentation Loss} The final predicted semantic mask $M$ is optimized using the coarse mask $M^c$ following
\begin{equation}
\label{eq:loss_CE}
    \mathcal{L}_S = -\dfrac{1}{N}\sum_{i}^{N}\sum_{j=0}^{n_c}\mathbbm{1}_j\left(M^c_{i}\right)\log\left(\sigma\left(M_{i}\right)_j\right)%
\end{equation}
where $\mathbbm{1}_j(x)$ is the indicator function and $\sigma (\mathbf{x})_j={\frac {\exp\left({x_j}\right)}{\sum_{k}^{n_c}\exp\left({x_k}\right)}}, \text{for} j\in\mathcal{C}$ is the softmax function.

\paragraph{Attention Loss} Inspired by the previous work where the box-shaped masks are used to train attention masks $\alpha$, the segmentation head is modified to include an identical and parallel convolution layer (\textcolor{attention}{yellow block} in Figure~\ref{fig:MTL-B4M}) to the last layer. Therefore, the attention map $\alpha=\left[\alpha^j\right]_{j\in\mathcal{C}}\in\mathbb{R}^{(n_c+1)\times N}$ shares the same dimension with the logit map $M^l$ and $\alpha^j\in\mathbb{R}^N, j\in\mathcal{C}$ is the attention map for class $j$. The predicted semantic mask $M$ is obtained by modulating the logits as $M = M^l \otimes \alpha$, where $\otimes$ is the Hadamard product. The attention map is optimized using the mean-squared error as shown in Eq.~\ref{eq:loss_MSE}.

\begin{equation}
\label{eq:loss_MSE}
    \mathcal{L}_\alpha = \dfrac{1}{N}\sum_i^{N}\left\|M^b_i - \alpha_i\right\|^2.
\end{equation}

\paragraph{Triplet Loss} To enforce consistency between predicted representations of un-annotated
    images, a triplet loss is proposed between the currently predicted ones, called queries,
    and those with annotations from previous batches, called keys.

    Let $n_B$ be the number of ground truth boxes of an image,
    $\mathcal{B}_k \subset \mathbb{N}^2$ be the \textit{index set} of the
    pixels within a ground truth box $B_k$, $k\in\{0\dots n_B-1\}$, and $\mathcal{M}_k\subset
    \mathcal{B}_k$ the \textit{index set} of the pixels having the same class
    prediction with $B_k$, or $\text{argmax}\left(M_i\right)=c_k,\forall i\in\mathcal{M}_k$.
    The unit-length mean embedding of the features vectors of the correctly
    predicted pixels in the bounding box $B_k$ is given by
    $\bar{\mathbf{z}}_k = \dfrac{1}{|\mathcal{M}_k|}\sum_{j\in\mathcal{M}_k} \mathbf{z}_j$,
    where $\|\bar{\mathbf{z}}_k\|=1$ and
    $\mathbf{z}$ is the feature vector input to the attention and logit layers (Figure~\ref{fig:MTL-B4M}).

    The triplet loss of a 3-tuple
    $\left(\bar{\mathbf{z}}_k,
    \bar{\mathbf{z}}^+_k,
    \bar{\mathbf{z}}^-_k\right)$
    is defined as
    
\begin{equation}
    \mathcal{L}_{\text{object}} = \sum_{k\in{0\dots n_B}}\max\left(0, \gamma +
    d_\mathbb{E}\left(\bar{\mathbf{z}}_k, \bar{\mathbf{z}}_k^+\right) -
    d_\mathbb{E}\left(\bar{\mathbf{z}}_k, \bar{\mathbf{z}}_k^-\right)\right),
\end{equation}

\noindent where $\bar{\mathbf{z}}_k^+$ be the embeddings of the same class $c_k$ with $\mathbf{z}$ and $\bar{\mathbf{z}}_k^-$ be the embeddings of different class, $d_\mathbb{E}$ is the Euclidean distance and  $\gamma$ is the margin, which is set to 0.1
The Box-for-Mask loss is the sum of all three loses, i.e. $\mathcal{L}_{B4M} = \mathcal{L}_S + \mathcal{L}_\alpha + \mathcal{L}_\text{object}$.

\section{Experiments}
\label{sec:exp}

\subsection{Setup}

\paragraph{Datasets.}
Most of the experiments are conducted on the Pascal VOC~\cite{PascalVOC} with extra ground truth semantic segmentation from the SBD dataset~\cite{Hariharan2011SBD} following the common practices. To simulate the partially-annotated scenario, we divide the training set into two halves, \hleft\ 7,558 images for object detection and \hright\ 7,656 for semantic segmentation following~\cite{Le2023BMVC} and report results on the 1,443-image validation set. The COCO dataset~\cite{mscoco} is used for the final result report. To simulate data scarcity, one-eighth of the training set (14,655 images) is uniformly sampled for training the detection task and another one-eighth (14,656 images) for semantic segmentation. The results are reported on the provided validation set of 5,000 images.

\paragraph{Implementation.} For backbone architectures, the ResNet (-18 and -50) and the SwinTransformer~\cite{Liu2021Swinv2} (-T, -B, -L) families are used. The FPN neck~\cite{Lin2017} is used with ResNet18 following~\cite{Le2023BMVC}, and PAFPN neck~\cite{Liu2018PAFPN} for the rest. The networks are trained for 70 epochs using SGD with a learning rate of $1\mathrm{e}{-3}$. We use the implementation of \cite{Le2023BMVC} which includes an EMA-updated network to collect trained parameters from the one trained with SGD. The EMA network is also used for the unbiased teacher~\cite{Liu2021ubteacher} in Table~\ref{tab:M4B-kd} and the B4M triplet loss.

\begin{table}[t]
    \centering
    \begin{tabular}{@{}llcccc@{}}
\toprule
Training   &$\mathcal{L}_\text{M4B}$& ResNet18  & ResNet50  &  SwinT    & SwinB     \\ %
\midrule                        
Baseline   &                        &    50.959 &    56.286 &  54.522   &   59.986  \\ %
\midrule                        
GT Masks   &C                       &    49.611 &    54.843 &   54.236  &   58.380  \\ %
           &L                       &    50.168 & \bf56.588 &\bf55.296  &\bf59.473  \\ %
           &L+C                     & \bf50.617 &    54.540 &   54.546  &   58.567  \\ %
\midrule                        
M4B Refined& C                      &    50.396 &    55.187 &   55.625  &   60.036  \\ %
           & L                      & \bf52.101 & \bf56.486 &\bf56.392  &\bf61.351  \\ %
           &L+C                     &    48.031 &    51.787 &   56.283  &   60.508  \\ %
\bottomrule
    \end{tabular}
    \caption{%
    Performance of M4B losses, including a localization (L) and/or a classification loss (C), 
    when combining mask-annotated and supervised images (baseline).
    M4B refinement with only localization loss outperforms using
    directly GT Masks' circumscribed rectangles.
    }
    \label{tab:M4B-prelim}
\end{table}

\begin{table}[t]
    \centering
    \begin{tabular}{@{}lccccccccc@{}}
        \toprule
        ResNet50    &   &Cls\da  &Loc\da  &Both\da &Dupe\da &Bkg\da  &Miss\da &FP\da   &FN\da    \\
        \midrule                                                                                   %
        Baseline    &   &   2.66 &\bf5.27 &   0.69 &\bf0.24 &   1.61 &   6.69 &\bf9.09 &   12.38 \\
        M4B Refined & C &   2.82 &   6.05 &   0.74 &   0.25 &   1.42 &   7.63 &   9.29 &   13.46 \\
                    & L &   2.76 &   5.74 &\bf0.66 &   0.31 &\bf1.36 &\bf6.48 &   9.47 &\bf12.19 \\
                    &L+C&\bf2.55 &   7.28 &   0.68 &   0.27 &   1.64 &   7.79 &   11.91&   12.92 \\
        \bottomrule
    \end{tabular}
    \caption{TIDE analysis for 3 configurations (L, C, L+C) of training with M4B Refined boxes on ResNet50.
    C only does not really improve Cls score but L+C does yet with the cost of Loc. L configuration has
    the most improvement especially on Miss and FN scores.
    }
    \label{tab:M4B-tide}
\end{table}

\begin{table}[t]
    \centering
    \begin{tabular}{@{}llccccc@{}}
\toprule
                                 &         &           & ResNet18 & ResNet50 &  SwinT  &  SwinB  \\    
\midrule 
Baseline                         & \hleft  &           &\bf50.959 &    56.286&   54.522&\bf59.986\\ %
GT Masks                         &\pie{360}&           &   50.168  &\bf56.588&\bf55.296&   59.473\\ %
\midrule 
U-Teacher~\cite{Liu2021ubteacher}&\pie{360}&EMA        &   34.882 &    43.008&   49.726&   55.364\\ %
M4B Refined                      &\pie{360}&EMA        &\bf52.101 &    56.486&\bf56.392&\bf61.351\\ %
\midrule 
U-Teacher~\cite{Liu2021ubteacher}&\pie{360}&pre-trained&   51.464 &    57.408&   53.236&   57.930\\ %
M4B Refined                      &\pie{360}&pre-trained&\bf53.517 & \bf57.721&\bf56.534&\bf60.956\\ %
\bottomrule
    \end{tabular}
    \caption{Employing M4B refinement with unbiased teacher knowledge distillation~\cite{Liu2021ubteacher}.
    Using pre-trained networks brings improvement, which can be pushed further by M4B refinement, while
    directly using EMA predictions as targets produces harmful effects.
    }
    \label{tab:M4B-kd}
\end{table}

\paragraph{Evaluation and analysis.} 
For object detection, the mAP metric implemented by Detectron2~\cite{Wu2019detectron2}, which averages APs at multiple IOU thresholds in $[0.5,0.55,\dots,0.95]$, is used and, for semantic segmentation, the conventional IOU score~\cite{Jaccard1912}. For analyzing the error sources, the TIDE~\cite{tide} framework is used with AP50-detection results, which shows 6 error types, including
\textbf{Cls}, localizing correctly but not classifying,
\textbf{Loc}, the other way around,
\textbf{Both}, localizing and classifying incorrectly,
\textbf{Dupe}, overlapping with a higher-scoring detection,
\textbf{Bkg}, detecting background as foreground,
\textbf{Miss}, all undetected ground truths not already covered,
as well as false positive \textbf{FP} and false negative \textbf{FN}.

\subsection{Mask for Box}
\label{sec:det-only}

In this section, we confirm the benefit of M4B loss for object detection. We compare two cases, when the circumscribed rectangles of the ground masks are used directly as referenced boxes for training (GT Masks) and when they are refined using the M4B. The results with classification loss (C) and both localization and classification (L+C) for pseudo targets are also shown for the ablation study. The results are shown in Table~\ref{tab:M4B-prelim}. It can be seen that using directly the ground truth masks' circumscribed rectangles would improve over the baseline due to more information being used and using M4B-refined boxes furthers the performance. Surprisingly, involving the classification loss when training with pseudo-targets (C and L+C) results in sub-optimal performance given that the pseudo-boxes' categories are correct by definition. The setting with only localization loss (L) consistently attains high performance. The TIDE analysis for ResNet50 in Table~\ref{tab:M4B-tide} shows that although L+C losses reduce Cls error,  the pseudo-boxes, generally, do not help improve accuracy (Cls, Loc, Dupe), yet training with pseudo-boxes for localization only is beneficial for FN (Bkg, Miss, FN) errors.

In Table~\ref{tab:M4B-kd}, M4B refinement is used with knowledge distillation, in particular, the Unbiased Teacher~\cite{Liu2021ubteacher} idea where the predictions, either from the EMA or a pre-trained network, are used as targets to train the current network. Without deviating input augmentation for teacher and student, which over-complicates the training pipeline, the same input images are used for both teacher (EMA/pre-trained) and student (SGD). Using EMA teacher predictions directly results in negative performance due to noises in the target boxes, especially during the first epochs. The pre-trained networks, on the other hand, improve performance as expected with knowledge distillation. M4B-refined runs are generally higher than their counterparts showing the accuracy of the refined boxes.

\subsection{Box for Mask}

\begin{table}[t]
  \centering
  \begin{tabular}{@{}llcccc@{}}
     \toprule
                                                      &         & ResNet18 & ResNet50  &   SwinT  &   SwinB  \\ %
     \midrule
     Baseline                                         &\hleft   &   65.292 &    69.065 &   75.012 &   79.690 \\
     $\; + \mathcal{L}_S$                             &\pie{360}&   64.740 &    67.582 &   73.830 &   78.510 \\
     $\;\quad +\mathcal{L}_\alpha$                    &\pie{360}&   67.623 &    71.763 &   76.269 &   80.654 \\
     $\qquad+$ B2S Affinity~\cite{Kulharia2020Box2Seg}&\pie{360}&   66.954 &    71.562 &   76.266 &   80.880 \\
     $\qquad+$ Shifting Rate~\cite{Song2023BCM}       &\pie{360}&   67.325 &    71.164 &   76.657 &\bf81.264 \\
     $\qquad+ \mathcal{L}_\text{object}$              &\pie{360}&\bf68.545 &\bf 72.306 &\bf76.981 &   81.236 \\
     \bottomrule
  \end{tabular}
  \caption{Performance of B4M losses when training box-annotated with supervised images (baseline).
  Simply applying cross-entropy loss ($\mathcal{L}_S$) on pseudo-masks (without attention maps)
  produces negative effects %
  but adding the attention loss ($\mathcal{L}_\alpha$) provides a boost. The triplet object
  loss ($\mathcal{L}_\text{object}$) is generally more helpful than the other approaches.
  }
  \label{tab:B4M}
\end{table}

This section validates the design choices for B4M losses. Pseudo-masks generated by GrabCut~\cite{GrabCut} are used for coarse pseudo-masks $M^c$. Table~\ref{tab:B4M} shows the performance when combining box-annotated with mask-annotated images (baseline) for training semantic segmentation. It can be seen that simply using cross-entropy loss with the pseudo-masks results in negative effects,  even with more training data, and half of them are fully annotated. The pseudo-masks, even when confined to only areas within respective ground truth boxes, are noisy and confuse the training process when wrong pixels are used as targets. The attention maps, trained with the box-shaped masks $M^b$, soften cross-entropy loss' strong imposition on all pixels, thus helping to take advantage of extra training data. Compared to Shifting Rate~\cite{Song2019BCM}, which requires fixed statistics pre-computed for each dataset and pseudo-labels, and comparing to Affinity loss~\cite{Kulharia2020Box2Seg}, the triplet object loss $\mathcal{L}_\text{object}$ improves performances.

\subsection{Combining Box-for-Mask and Mask-for-Box}

In this section, we combine the two previous modules, \underline{Bo}x-for-Mask and \underline{M}ask-for-\underline{Bo}x, or BoMBo, in one network for multi-task partially supervised learning. We choose $\lambda=2$ in Eq.~\ref{eq:loss} to balance detection and segmentation losses. Table~\ref{tab:BoMBo-VOC} shows the results with and without BoMBo on the VOC and COCO dataset. Although BoMBo outperforms all baselines on COCO, it only excels on both tasks for SwinL on VOC and has mixed results on the other architectures. The results show the benefit of the weak losses in using the extra data available to train one task with the other's annotations but also suggest an imbalance problem when combining training signals for both tasks.

\begin{table}[t]
    \centering
    \setlength{\tabcolsep}{4pt}
    \begin{tabular}{@{}lccccc@{}}
        \toprule
        Training          &\multicolumn{4}{c}{Detection}                  \\
        \cmidrule{2-5}
                          & ResNet50 & SwinT     & SwinB     & SwinL     \\ %
        \midrule                                  
        VOC: MTL          &\bf 55.174&    53.305 & \bf58.267 &   59.713  \\ %
        + BoMBo           &    54.885& \bf54.696 &    58.259 &\bf60.687  \\ %
        \midrule                                  
        COCO: MTL         &    17.198&    15.158 &    14.914 &   19.766  \\ %
        + BoMBo           &\bf 19.087& \bf16.918 & \bf17.416 &\bf21.935  \\ %
        \bottomrule
    \end{tabular}~
    \begin{tabular}{@{}cccc@{}}
        \toprule
        \multicolumn{4}{c}{Segmentation}                    \\
        \cmidrule{1-4}       
                           ResNet50  &SwinT      & SwinB     &SwinL      \\ %
        \midrule                                            
                           \bf75.658 & \bf77.795 &\bf 81.798 &    83.093 \\ %
                              74.861 &    77.433 &    81.205 &\bf 83.310 \\ %
        \midrule                                            
                              54.535 &    56.280 &    63.802 &    67.788 \\ %
                           \bf58.466 & \bf59.102 &\bf 66.420 &\bf 68.968 \\ %
        \bottomrule
    \end{tabular}
    \caption{BoMBo on VOC and COCO datasets.
    }
    \label{tab:BoMBo-VOC}
\end{table}

\section{Discussion and Conclusion}
\label{sec:conclusion}

The paper investigates various weak losses for training object detection from ground truth semantic masks and vice versa semantic segmentation from box annotations in multi-task partially supervised learning. To that end, the two modules, Box for Mask and Mask for Box and their combination BoMBo, are proposed. The ablation studies show that naïvely extracting shared information to train the other task might result in negative impacts even when supervised data are also being used. The pseudo-boxes, despite having correct categories, can only help when being trained only for localization while pseudo-semantic masks should only constrain the attention-modulated predictions. One limitation of the study is the assumptions of the same data domain and shared class space between the two tasks, which needs to be addressed if partially annotated data are to be used to expand the training set.

\section*{Acknowledgments}
This work was supported by the SAD 2021 ROMMEO project (ID 21007759) and the ANR AI chair OTTOPIA project (ANR-20-CHIA-0030).

\bibliography{macro,IRISA-BMVC24}

\newpage

\section{Network architecture}

\begin{figure}
    \centering
    \def\svgwidth{\textwidth}
    \scriptsize
    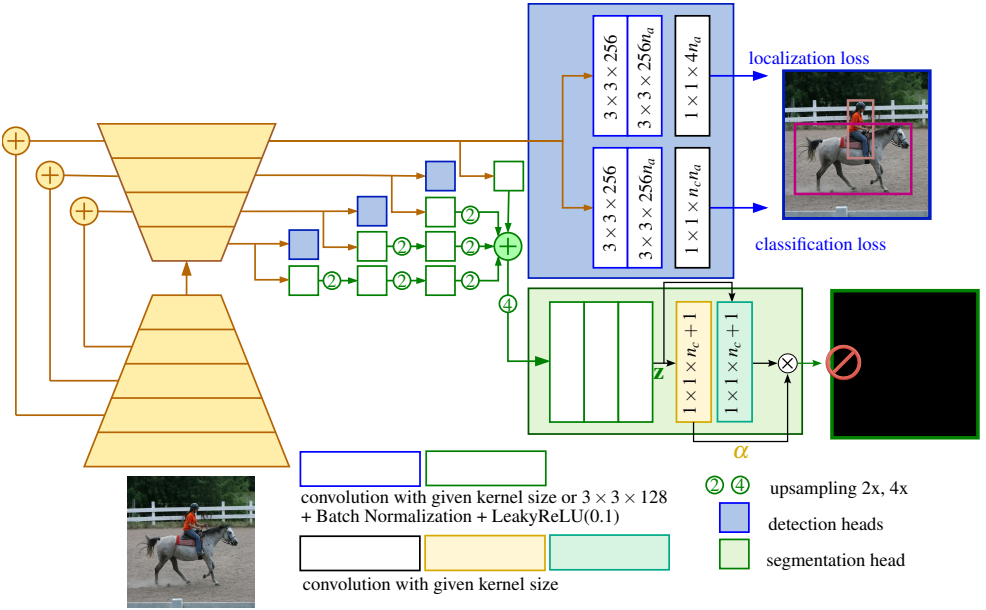
    \caption{The network architecture being used in the paper, redrawn from~\protect\cite{Le2023BMVC} with an additional attention module in the segmentation head. The figure is illustrated with a detection-annotated input, thus cannot train the semantic segmentation head.
    }
    \label{fig:full_MTL}
\end{figure}

Figure~\ref{fig:full_MTL} presents the full network, reused from~\cite{Le2023BMVC}, which comprises a backbone and FPN neck as the encoder, a detection head at each FPN level, and a segmentation head using aggregated features from a FPN panoptic~\cite{Kirillov2019panoptic} subnet. In general, the network architecture is kept unchanged from~\cite{Le2023BMVC} except an additional attention module in the segmentation head.

\section{Mask-for-Box module}

Algorithm~\ref{alg:M4B} shows in detail the matching and refining operations for the Mask-for-Box module.
The algorithm receives as inputs the ground truth semantic segmentation mask and a list of predicted boxes $\hat{B}=\left\{\left(\hat{L}_k,\hat{c}_k\right)\right\}^N_{k=0}$ where $N=w\times h$ is spatial dimension of the feature output at each FPN scale,$\hat{L}_k\in\mathbb{R}^4$ is a box localization containing 4 coordinates of a box, and $\hat{c}_k\in\mathbb{R}^{n_C}$ is the vector of confidence scores for each of the $n_C$ category output by a sigmoid function.
The \texttt{touch} function (L13, 31) makes sure that a predicted box $\hat{L}_k$ touches at least two sides of a reference box $L$ (by comparing to a distance of 10\% of the reference box's width and height). The \texttt{crop} function (L14,32) creates a box with the reference box's touching side and the non-touching side of the predicted box. The powerset $\mathcal{P}$ (L21) returns the set of all subsets of $\tilde{B}^j$ while the \texttt{merge}$(p)$ function creates the smallest box covering all the reference boxes in $p$. The \texttt{unique} function makes sure the same merged box is processed only once. The leftover variable contains the referenced boxes to be processed, it first receives all referenced boxes (L10) then has the boxes removed once they are decided to be either in the splitting (L18) or merging case (L26). The shortlist variable keeps all the predicted boxes matched to the referenced ones which is filtered by the non-maximum impression \texttt{NMS} function (L34). The referenced boxes not recognized for splitting nor merging are added to the output as is.

\renewcommand{\algorithmicthen}{}
\begin{algorithm}[t]
    \caption{Mask-for-Box matching and refining}
    \begin{algorithmic}[1]
        \renewcommand{\algorithmicrequire}{\textbf{Input:}}
        \renewcommand{\algorithmicensure}{\textbf{Output:}}
        \REQUIRE predicted boxes  $\hat{B} = \left\{\left(\hat{L}_k,\hat{c}_k\right)\right\}^N_{k=0}, \hat{L}_k\in\mathbb{R}^4, \hat{c}_k\in\mathbb{R}^{n_C}$, ground truth masks
        \ENSURE refined boxes
        \STATE references = $\left\{\right\}$ \COMMENT {obtained from the ground truth masks}
        \FOR{each unique category $j$ in a ground truth mask}
           \STATE extract all connected components of this category
           \FOR {each connected component}
               \STATE $L\leftarrow$ extract circumscribed rectangle
               \STATE references $\overset{+}{\leftarrow} (L, j)$
            \ENDFOR
        \ENDFOR
        \STATE leftover = $\left\{\right\}$
        \STATE shortlist = $\left\{\right\}$ \COMMENT {A. Splitting case}
        \FOR {each box $\tilde{b} = \left(L, j\right)$ in references}
            \STATE leftover $\overset{+}{\leftarrow} \tilde{b}$
            \FOR {each predicted box $\hat{b}_k = \left(\hat{L}_k,\hat{c}_k\right)\in\hat{B}, \hat{c}_k^j > 0.4$}
                \IF {\texttt{IOU}$\left(L, \hat{L}_k\right) \leq 0.6$}
                    \IF {\texttt{touch}$\left(L, \hat{L}_k\right)$}
                        \STATE $L \leftarrow \texttt{crop}(L,\hat{L}_k)$
                    \ELSE
                        \STATE \texttt{continue} \COMMENT{skip this predicted box}
                    \ENDIF
                \ENDIF
                \STATE shortlist $\overset{+}{\leftarrow}\left(L,\min\left(0.9, \hat{c}_k^j + 0.1\right)\right)$
                \STATE leftover $\overset{-\tilde{b}}{\rightarrow}$ %
            \ENDFOR
        \ENDFOR
    
        \COMMENT {B. Merging case}
        \FOR {each unique category $j$ from the left-over references}
            \STATE $\tilde{B}^j\leftarrow$  all leftover boxes of category $j$
            \STATE $\tilde{B}\leftarrow$ \texttt{unique}$\left(\texttt{merge}\left(p\right),\forall p\in\mathcal{P}\left(\tilde{B}^j\right), \|p\| > 1\right)$
            \FOR {each box $\tilde{b} = (L, j)\in\tilde{B}$}
                \STATE $\left(\hat{L}_\text{IOU},\hat{c}_\text{IOU}\right) \leftarrow \text{argmax}_{\hat{b}_k}\texttt{IOU}\left(L, \hat{L}_k\right),\forall \hat{b}_k\in\hat{B}$ 
                \IF {$\hat{c}_k^j > 0.1$}
                    \STATE shortlist $\overset{+}{\leftarrow}\left(L,\min\left(0.9, \hat{c}^j_\text{IOU}+0.4\right)\right)$
                    \STATE leftover $\overset{-p}{\rightarrow}$
                \ENDIF
            \ENDFOR
        \ENDFOR
        
        \COMMENT{C. Adding well-predicted boxes}
        \FOR {each box $\tilde{b} = \left(L, j\right)$ in references}
            \FOR {each predicted box $\hat{b}_k = \left(\hat{L}_k,\hat{c}_k\right)\in\hat{B}, \hat{c}_k^j > 0.5$}
                \IF {\texttt{IOU}$\left(L, \hat{L}_k\right) \geq 0.8$}
                    \STATE shortlist $\overset{+}{\leftarrow}\left(L,\hat{c}^j_k\right)$
                \ELSIF{$\texttt{touch}(L, \hat{L}_k)$}
                        \STATE $L\leftarrow\texttt{crop}(L,\hat{L}_k)$
                        \STATE shortlist $\overset{+}{\leftarrow}\left(L,\hat{c}^j_k\right)$
                \ENDIF
            \ENDFOR
        \ENDFOR

        \STATE refined $\leftarrow$ \texttt{NMS}(shortlist, 0.4)
        \STATE refined $\overset{+}{\leftarrow}$ leftover \COMMENT {D. Left over boxes}
    \end{algorithmic}
    \label{alg:M4B}
\end{algorithm}

\section{Qualitative results}

\begin{figure}[t]
    \centering
    \includegraphics[width=\linewidth]{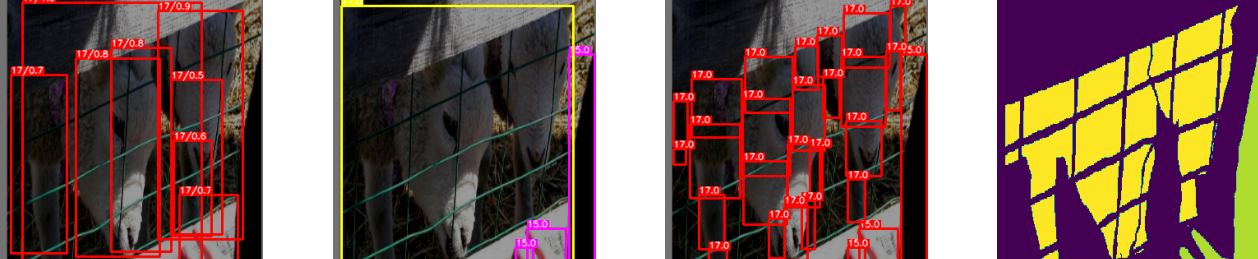} %
    \includegraphics[width=\linewidth]{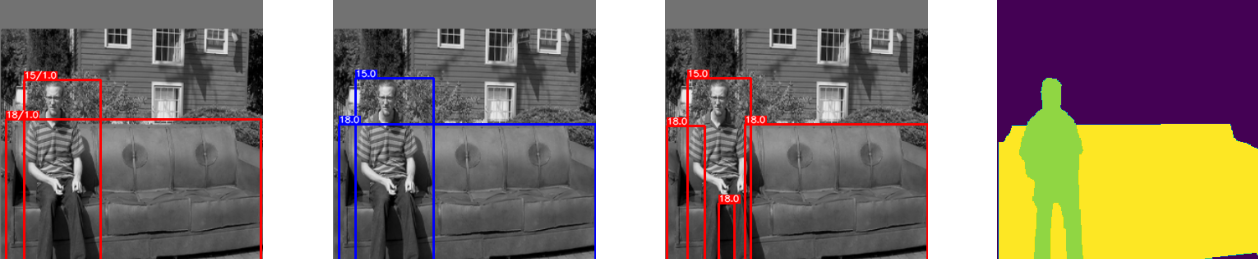} %
    \includegraphics[width=\linewidth]{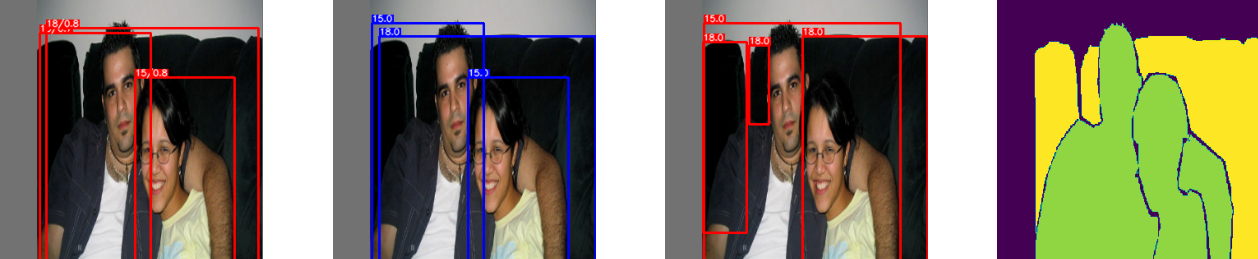} %
    \includegraphics[width=\linewidth]{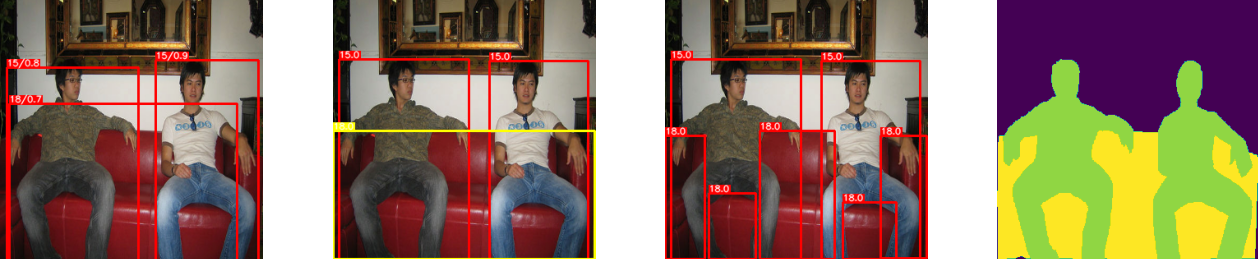} %
    \includegraphics[width=\linewidth]{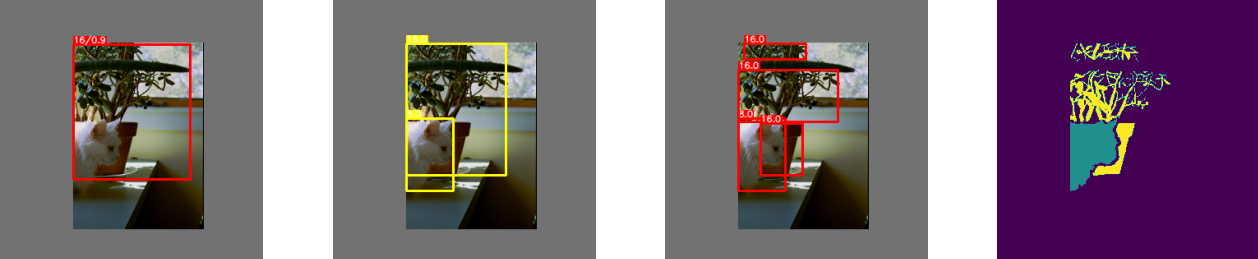} %
    \caption{Qualitative results, from left to right: predicted boxes, refined boxes by M4B, boxes from ground truth masks, ground truth masks. The circumscribed boxes of the ground truth follow strictly the connected components, hence can include multiple objects or be fragmented. The refined boxes use the learned information from the predicted box and categories from the masks to obtain better-fitted boxes.}
    \label{fig:qualitative1}
\end{figure}

\begin{figure}[t]
    \centering
    \includegraphics[width=\linewidth]{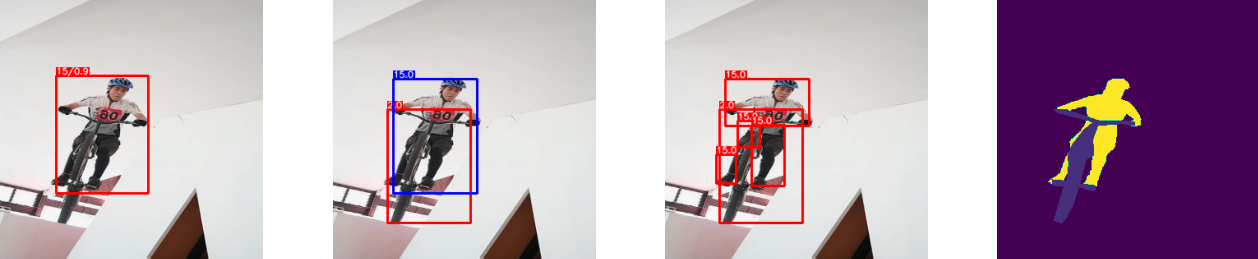} %
    \includegraphics[width=\linewidth]{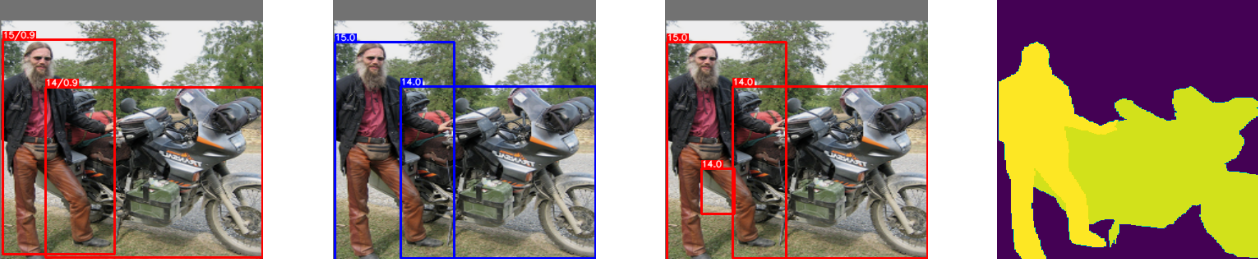} %
    \includegraphics[width=\linewidth]{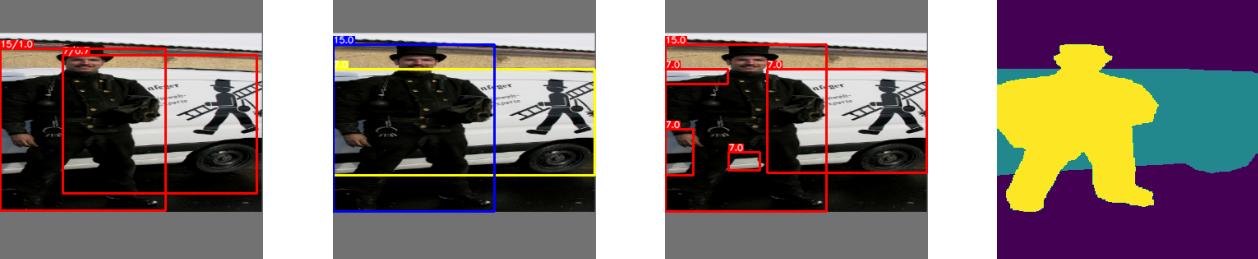} %
    \includegraphics[width=\linewidth]{Images/m4b_illust/2_feh_2757020_000010_epoch-61.png} %
    \includegraphics[width=\linewidth]{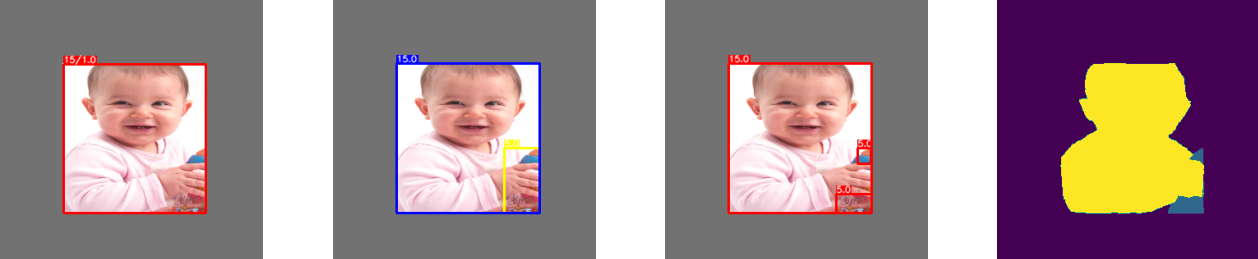} %
    \caption{Qualitative results, from left to right: predicted boxes, refined boxes by M4B, boxes from ground truth masks, ground truth masks. The circumscribed boxes of the ground truth follow strictly the connected components, hence can include multiple objects or be fragmented. The refined boxes use the learned information from the predicted box and categories from the masks to obtain better-fitted boxes.}
    \label{fig:qualitative2}
\end{figure}

\begin{figure}
    \centering
    \includegraphics[width=\linewidth]{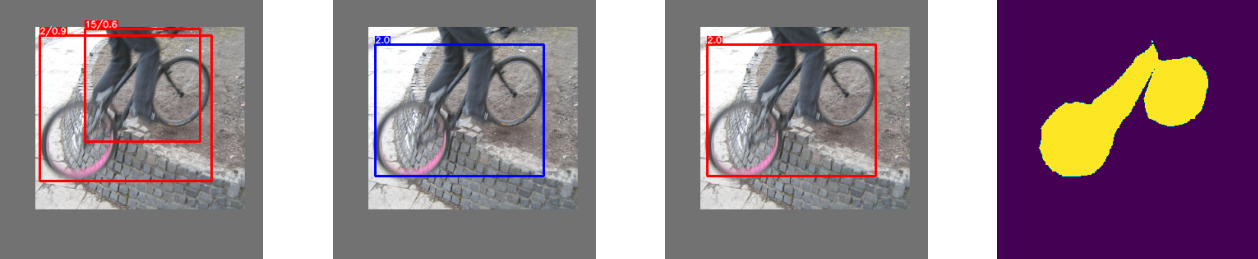} %
    \includegraphics[width=\linewidth]{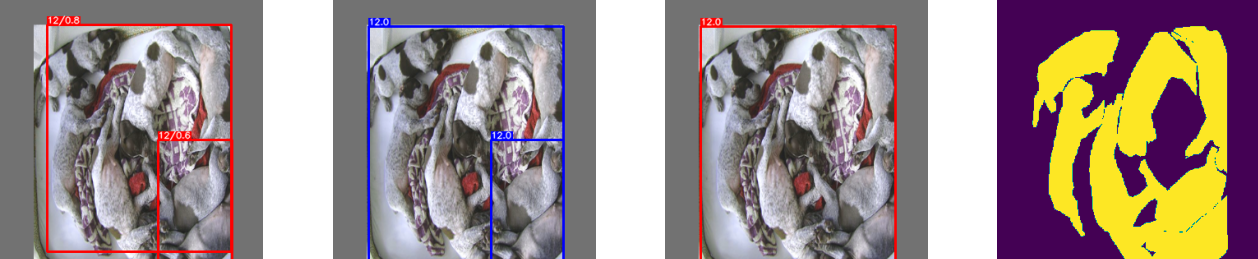} %
    \includegraphics[width=\linewidth]{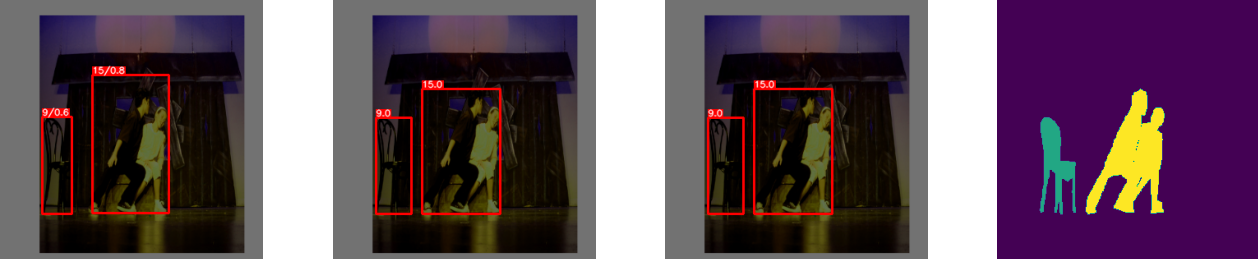} %
    \includegraphics[width=\linewidth]{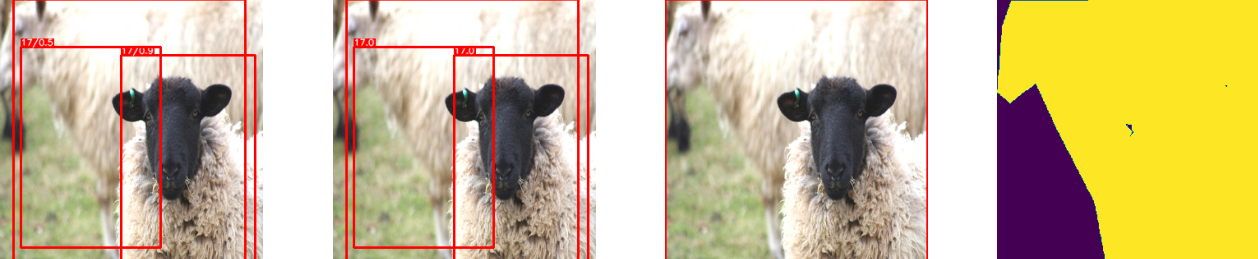} %
    \includegraphics[width=\linewidth]{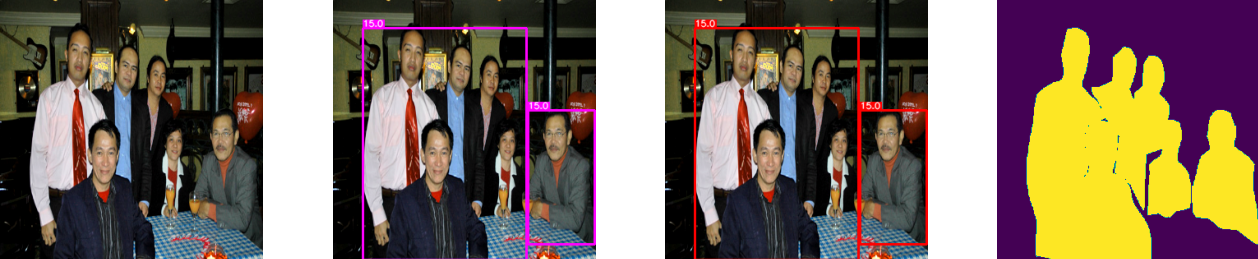} %
    \includegraphics[width=\linewidth]{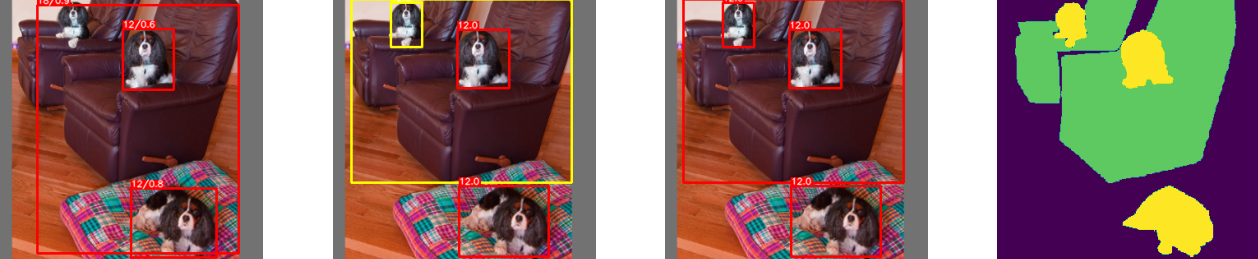} %
    \caption{Qualitative results of failure cases, from left to right: predicted boxes, refined boxes by M4B, boxes from ground truth masks, ground truth masks. The refined boxes are incorrect when there is not sufficient information from both sources.}
    \label{fig:failure}
\end{figure}

Figure~\ref{fig:qualitative1} and~\ref{fig:qualitative2} illustrate the results of refined boxes from the M4B module. The refined boxes use the best of both worlds, the instance information from the predicted boxes and class information from the ground truth masks to add missing boxes, separate multi-instance boxes, and merge fragmented ones. Some failure cases are shown in Figure~\ref{fig:failure}.

\end{document}